\documentclass[lettersize,journal]{IEEEtran}
\usepackage{amsmath,amsfonts}
\usepackage[linesnumbered,ruled,vlined]{algorithm2e}
\usepackage{array}
\usepackage{textcomp}
\usepackage{stfloats}
\usepackage{url}
\usepackage{verbatim}
\usepackage{graphicx}
\usepackage{cite}
\usepackage{amssymb}
\usepackage{mathtools}
\usepackage{amsthm}
\usepackage{enumitem}
\usepackage{microtype}
\usepackage{graphicx}
\usepackage{subcaption}
\usepackage{booktabs}
\usepackage{tcolorbox}
\usepackage{multirow}
\usepackage{color}
\usepackage{bm}
\usepackage{newfloat}
\usepackage{listings}
\tcbuselibrary{listings, breakable}

\begin{document}

\title{Evolving Pixel-Based Game Agent: \\ Generalizable Strategy Learning from Open-World Exploration\thanks{Corresponding author: Chenwei~Tang}
\thanks{X. Liu, C. Tang and J. Lv are with the College of Computer Science, Sichuan University, and Engineering Research Center of Machine Learning and Industry Intelligence, Ministry of Education, Chengdu 610065, PR China (e-mail: xinyuliu0804@stu.scu.edu.cn, \{tangchenwei, lvjiancheng\}@scu.edu.cn).}
\thanks{Z. Wang, J. Du and L. Zhen are with the Institute of High Performance Computing, Agency for Science, Technology and Research (A*STAR), Singapore 138632, Singapore}
}

\author{
Xinyu Liu,
Chenwei Tang,
Zizhou Wang, 
Jiawei Du, 
Liangli Zhen,
and Jiancheng Lv
}

\maketitle

\begin{abstract}
Enabling agents to autonomously master complex, open-ended environments with minimal human preparation is a critical step toward general-purpose agents. However, existing approaches often rely on labor-intensive task-specific priors, e.g., manually annotating interactive elements, predefining action spaces, or engineering rewards and workflows, which become brittle as game rules, units, or UI evolve over time, especially in strategy games with large, dynamic state spaces. To eliminate this dependency, we propose the \textbf{Evolving Pixel-Based Game Agent} (\textbf{EPGA}), a framework that enables autonomous learning in complex, open-ended strategy games directly from raw pixels. EPGA continuously builds and updates its understanding of the game by discovering interactive elements, maintaining contextual and experiential memory, tracking long-horizon game progress, and adaptively refining its action strategy. These capabilities are realized through four synergistic components: structured memory for contextual recall and experience generalization, continual object discovery for identifying novel interactive elements, confidence-guided action updating for balancing exploration and exploitation, and game-process tracking for phase-aware planning over extended horizons. Experiments on two challenging strategy games, i.e., \textit{Slay the Spire} and \textit{Civilization V}, demonstrate that EPGA consistently outperforms strong baselines that rely on manually curated task-specific priors, achieving a top in-game score of 109 in \textit{Slay the Spire}, while maintaining an execution-responsive rate above 98.73\%. Moreover, EPGA exhibits robust adaptation to unseen games and sustained learning in non-stationary environments, demonstrating its potential as a scalable paradigm for long-horizon autonomous agents.
\end{abstract}


\section{Introduction}
\label{sec:intro}
Autonomous learning in complex strategy games, and more broadly, dynamic decision-making environments, remains a fundamental challenge for developing real-world capable agents. This challenge stems primarily from three core environmental properties \cite{guo2023rethinking, fang2024large}. First, the presence of \textit{dynamically evolving state spaces} \cite{zhou2024hazard,xu2024survey,chen2024llmarena} in environments requires agents to continuously detect and adapt to newly emerging interactive elements. Second, \textit{long-term planning horizons} \cite{zhang2024fltrnn, wang2025memory,bai2025maple} demand consistent strategy tracking and phase-aware reasoning over extended trajectories. Third, \textit{combinatorial action complexity} \cite{du2025rethinking,bansal2017emergent,dockhorn2021game} arises from structured, high-dimensional decision spaces with intertwined dependencies. These characteristics critically limit conventional learning paradigms: Traditional Reinforcement Learning (RL) \cite{pang2019reinforcement,xie2026reinforcement,uehara2026review} pipelines often fail when environmental shifts invalidate predefined state or reward definitions, necessitating costly retraining. Similarly, LLM-based agents \cite{li2024survey,chowa2026language,piccialli2025agentai} are prone to cascading reasoning failures when confronted with unseen game elements or changes in actionable space, due to their reliance on handcrafted priors and static tool APIs \cite{hong2023metagpt, park2023generative, wang2023voyager, bonatti2024windows}. 

\begin{figure}
    \centering
    \begin{subfigure}{\columnwidth}
        \centering 
        \includegraphics[width=\linewidth]{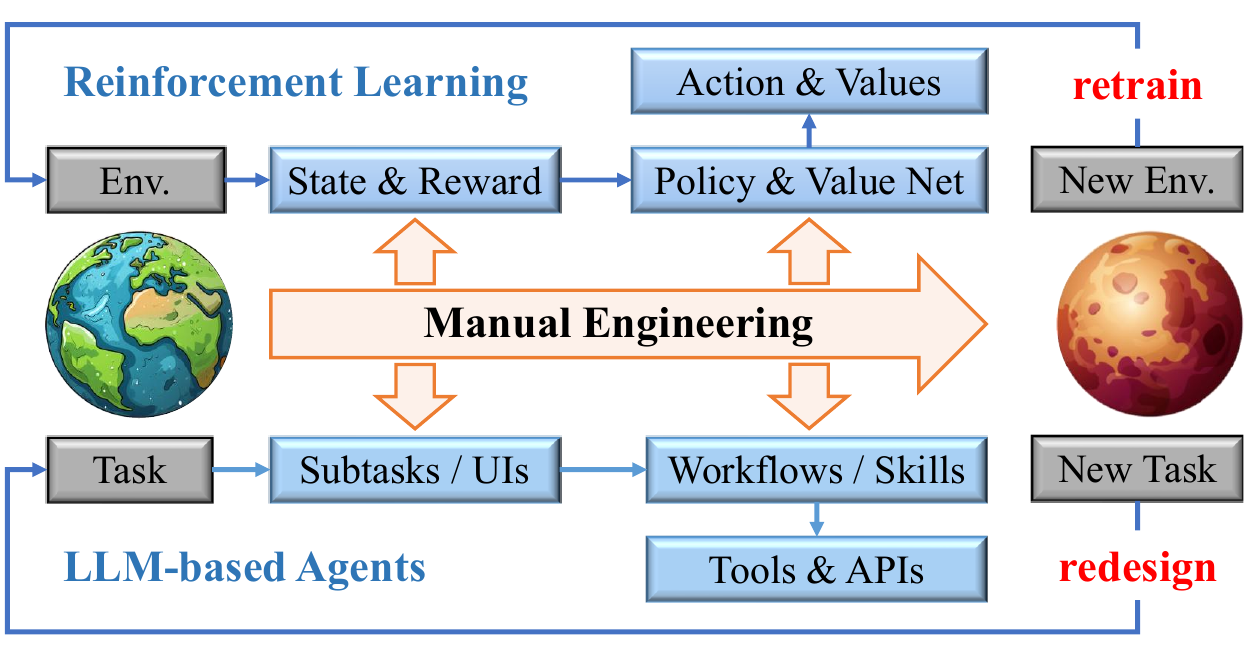}
        \caption{Mainstream Paradigms: RL \& LLM-based Agents}
        \label{fig:moti1}
    \end{subfigure}
    \hfill
    \begin{subfigure}{\columnwidth}
        \centering
        \includegraphics[width=\linewidth]{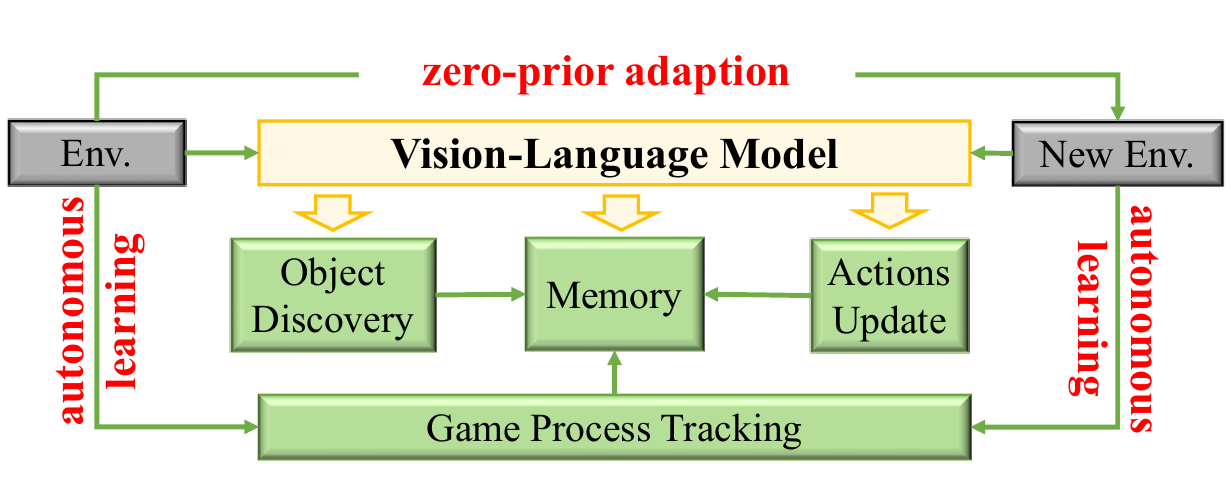}
        \caption{Our Evolving Pixel-Based Game Agent}
        \label{fig:moti2}
    \end{subfigure}
\caption{Comparison of conventional agent paradigms and \textbf{EPGA} for adapting to evolving environments. Traditional RL methods and LLM-based agents rely on manually designed state representations, action spaces, rewards, task decompositions, and interactive tools, requiring substantial redesign when adapting to new tasks. In contrast, EPGA starts from raw pixels and empty memory, autonomously discovering and verifying novel interactive objects, enabling continual adaptation without human intervention.}

    \label{fig:moti}
\end{figure}

As shown in Figure~\ref{fig:moti1}, conventional agent design paradigms face critical limitations in complex strategy game environments. Traditional RL methods require laborious manual engineering to define state/action spaces and reward functions \cite{boggess2025explanations, qu2025latent, tang2025deep}, creating a rigid dependency between policy networks and predefined environmental setups. This coupling often forces full retraining when environments change, as existing parameterizations become obsolete. In contrast, LLM-based workflow agents rely on static task decompositions and predefined toolchains \cite{qian2023chatdev, agarwal2024multi, niu2025flow}, executing reasoning through fixed APIs or scripted procedures. Even minor environmental shifts can trigger cascading failures across these brittle pipelines due to error propagation along sequential subtasks. More recently, multimodal agents built on large Vision-Language Models (VLM) \cite{du2025rethinking,tan2024cradle,tang2025experience}, e.g., CRADLE, Bottom-Up Agent, SAG-Agent, have shown promise in end-to-end decision-making by integrating visual and linguistic reasoning. Yet, they remain limited by predefined UI interaction patterns and inadequate long-term temporal modeling, often resulting in short-sighted and inflexible strategies when game mechanics or interfaces evolve. Together, these limitations highlight the urgent need for agents capable of continuous learning and robust adaptation in non-stationary, open-world environments.

To this end, we propose \textbf{Evolving Pixel-Based Game Agent} (\textbf{EPGA}) to enable autonomous learning directly from raw pixels, as shown in Figure~\ref{fig:moti2}. Departing from prior RL and LLM agents that rely on predefined objects, actions, or APIs, the proposed \textbf{EPGA} features a unique action-space construction mechanism: it autonomously discovers interactable entities from pixels and dynamically builds actions during exploration. This design supports continuous and scalable expansion of state and action spaces without manual intervention. \textbf{EPGA} integrates four synergistic modules, including \textit{memory module}, \textit{object discovery module}, \textit{action update module}, and \textit{game process tracking module}. Specifically, the \textit{memory module} supports experience retention and contextual recall by storing structured traces of past interactions, including objects, action sequences, and Monte Carlo Tree Search (MCTS) outcomes \cite{swiechowski2023monte}, to facilitate generalization across similar situations. Next, the \textit{object discovery module} enables continual discovery by comparing and reasoning over visual inputs to identify previously unseen elements, thus expanding the agent’s interaction space in evolving environments. After that, the \textit{action update module} supports adaptive behavior by evaluating decisions based on observed feedback and updating confidence-weighted activation scores, allowing the agent to balance exploration and exploitation effectively. Finally, the \textit{game process tracking module} provides long-horizon monitoring by detecting phase transitions and encoding global context, thereby guiding temporally consistent planning and mitigating short-sighted behavior. Together, these four modules form a cohesive framework that systematically addresses the core challenges, e.g., self-directed exploration, decision-making, introspection, and progress tracking, of autonomous agents operating in complex strategy games. Our key contributions are as follows:

\begin{itemize}[leftmargin=*, nosep, topsep=8pt]
    \item We propose a prior-free paradigm that continuously discovers interface elements and constructs action spaces directly from pixels, eliminating the need for manual annotation or prompt engineering when encountering unseen game content.  
    \item The proposed \textbf{EPGA} progressively builds structured representations of interactive objects, executable skills, and game processes, enabling the agent to reason with expanding context and perform long-horizon planning.
    \item Extensive evaluations demonstrate that \textbf{EPGA} consistently outperforms strong baseline methods across two challenging game environments. These results validate the effectiveness of the proposed autonomous learning paradigm in both performance and zero-prior adaptation.
\end{itemize}

\section{Related Work}
\textbf{Reinforcement Learning.} Traditional RL methods, e.g., PPO \cite{schulman2017proximal} and SAC \cite{haarnoja2018soft}, rely heavily on manually specified state and action spaces. These predefined representations work well in stable environments, but when game mechanics change—such as altering state dimensions or introducing new interaction types—agents often require substantial modifications and full retraining. Additionally, reward functions are typically handcrafted and rarely generalize across tasks. Meta-RL methods like MAML \cite{duan2016rl} and RL$^2$ \cite{finn2017model} improve adaptability by training over task distributions, allowing agents to acquire general adaptation strategies across related tasks. However, they still depend on handcrafted state encodings and rewards, and assume distributional consistency between training and testing tasks, limiting applicability when dynamic environments introduce novel game mechanics or interaction patterns \cite{nakhaeinezhadfard2025entropy,liu2026mixture}.

\textbf{Agent AI.} LLM-based workflow agents, e.g., ReAct \cite{yao2023react} and AutoGPT \cite{yang2023auto}, achieve task execution via human-authored procedures and static toolchains. While effective in narrow domains, they exhibit three core limitations: \textit{static behavior}, \textit{reliance on handcrafted priors}, and \textit{short-term operational horizon}. In particular, their effectiveness often depends on predefined task decompositions, workflows, and available interaction tools, which limits their ability to respond to previously unseen situations. More critically, such agents lack adaptability in complex environments—like evolving strategy games—where game rules, UI layouts, or mechanics frequently change. Recent multimodal large models, such as GPT-4o \cite{openai2024gpt4o}, DeepSeek-R1 \cite{guo2025deepseek}, and Claude \cite{anthropic2025claude}, empower autonomous agents with strong visual understanding, language reasoning, and multi-hop planning abilities \cite{du2025rethinking, tan2024cradle,tang2025experience, cabannes2024iteration, lin2025showui}, enabling end-to-end learning of semantic states from raw visual and textual inputs. These capabilities make multimodal agents promising for strategy games that require integrated perception, reasoning, and planning. Nevertheless, they still face key limitations: they rely on predefined interactive UI elements, making them brittle to dynamically evolving interfaces, and they make decisions based only on the current frame without persistent long-term temporal memory. As a result, agents may fail to recognize and reuse newly discovered interactions, or produce short-sighted and inconsistent strategies when the environment changes over time.

\begin{figure*}
\centering
\includegraphics[width=\textwidth]{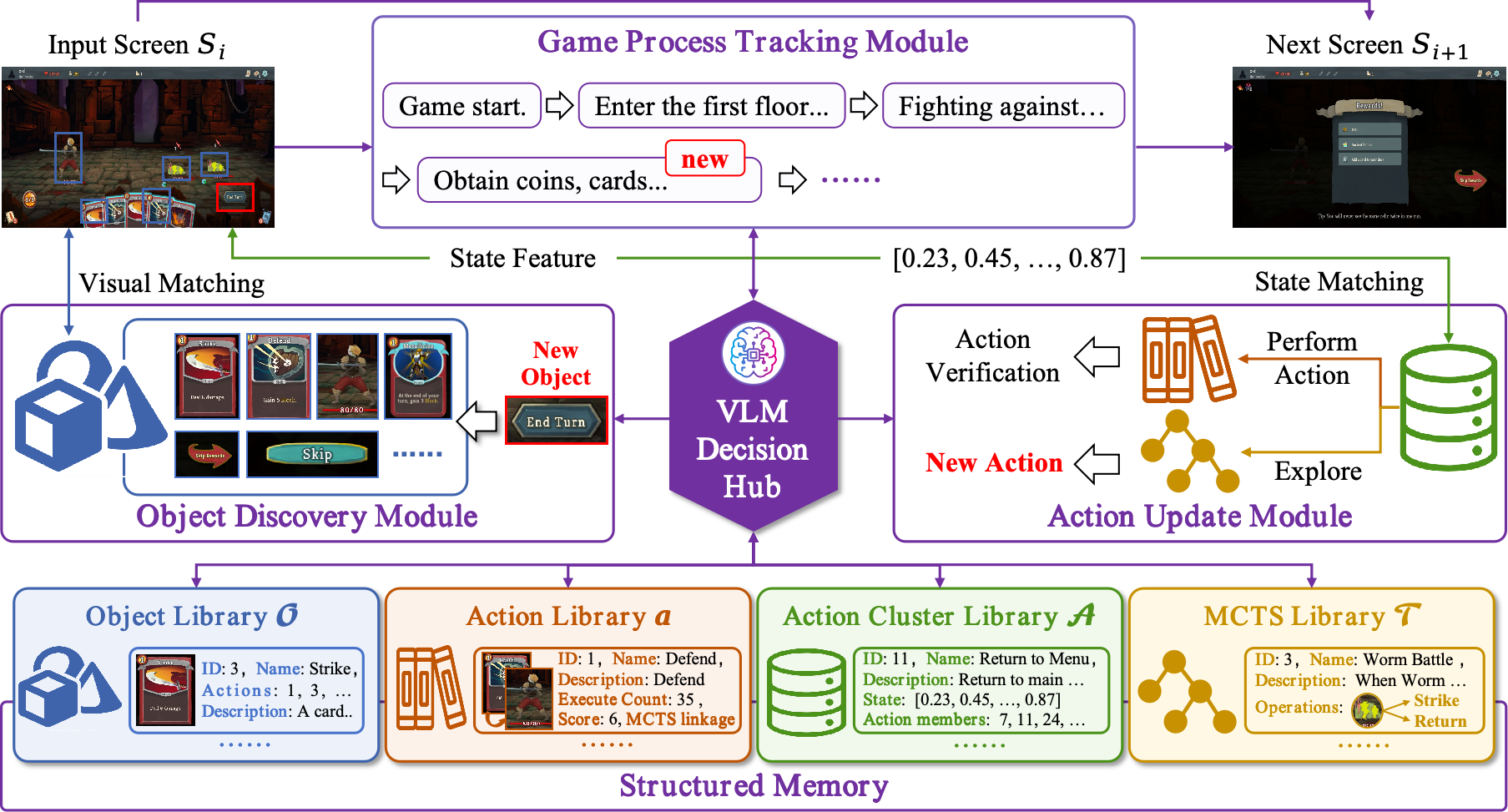}
\caption{Architecture of the \textbf{Evolving Pixel-Based Game Agent (EPGA)}. EPGA matches visual elements against structured memory and uses a VLM to discover novel interactable objects through the \textit{Object Discovery Module}. It retrieves context-relevant actions, executing high-confidence actions or exploring new ones via the \textit{Action Update Module}. Interaction outcomes are stored in \textit{Structured Memory} for experience generalization, while the \textit{Game Process Tracking Module} detects high-level phase shifts to support long-horizon planning. Together, these components enable autonomous policy learning and continual adaptation without human intervention.}
\label{fig:Our framework}
\end{figure*}

\section{Methodology}
As shown in Figure~\ref{fig:Our framework}, the \textbf{EPGA} integrates four core modules designed to support autonomous learning directly from raw pixels. To enable adaptation in dynamically evolving environments without relying on predefined action spaces or handcrafted features, we build upon VLMs \cite{li2025benchmark}, which provide scalable visual-linguistic reasoning for scene understanding and interaction guidance. In our framework, the VLM (e.g., GPT-4o) serves not merely as a perceptual tool but as a central component that drives object discovery, action verification, strategy generation, and progress tracking. As detailed in Table~\ref{tab:vlm_roles}, the VLM performs distinct functional roles across specialized sub-components, enabling continuous interpretation and adaptation within open-world game settings. In the following sections, we describe each module in turn, emphasizing its contribution to autonomous learning and decision-making. The section concludes with a discussion of design considerations that improve sample efficiency and prioritize high-value behaviors.

\subsection{Structured Memory Module}
To support long-term strategy learning in dynamic environments, agents must not only make immediate decisions, but also systematically retain and recall useful past experiences. Unlike the VLM, which provides semantic understanding to inform action selection, the memory module serves as the experience backbone of the system—accumulating, organizing, and supplying historical information to guide the continual updates of objects, actions, and game understanding. Our structured memory maintains four repositories: 1) \textit{\textbf{Object Library}} \bm{$\mathcal{O}$}: maintains a collection of verified objects represented as triplets $(O^v_i, O^n_i, O^a_i)$, where $O^v_i$ denotes the cropped visual instance, $O^n_i$ is the textual description of its function, and $O^a_i$ is the corresponding action associated with the object. 2) \textit{\textbf{Action Library}} \bm{$\mathit{a}$}: record constructed actions with associated metadata—execution count $n_i$, fitness score $f_i$, and Monte Carlo tree linkage—supporting policy refinement and sampling. 3) \textit{\textbf{Action Cluster Library}} \bm{$\mathcal{A}$}: organize semantically similar actions into clusters, each indexed by a representative state $S_i = (S^{scr}_i, S^{fea}_i)$ that includes both the screen image and its CLIP \cite{radford2021learning}-encoded feature, enabling state-aware matching and efficient retrieval. 4) \textit{\textbf{MCTS Tree Library}} \bm{$\mathcal{T}$}: maintain simulation trees used during planning, where nodes correspond to discovered objects and edges denote valid executable actions. Through this design, the memory module distills exploratory outcomes into structured, reusable knowledge—accelerating decision-making and supporting continual updates across all other modules.

\begin{table*}[t]
\centering
\caption{Functional roles with inputs, responsibilities, and outputs of the VLM in the \textbf{EPGA}. By decomposing its functionality into discrete roles, e.g., object discovery, action generation, and progress tracking, the VLM enables the agent to dynamically interpret, act, and adapt in open-world game environments without manual supervision or predefined action spaces.}
\begin{tabular}{|>{\raggedright\arraybackslash}m{2.9cm}|>{\raggedright\arraybackslash}m{5.2cm}|>{\raggedright\arraybackslash}m{3.7cm}|>{\raggedright\arraybackslash}m{4.2cm}|}
\hline
\textbf{Component} & \textbf{Input} & \textbf{Responsibility} & \textbf{Output} \\
\hline
\textbf{\textit{Object Observer}} & Current screen $S^{scr}_i$ + Process $P_i$ +\newline Known object list $\mathcal{O}_{exist}$ & Detect candidate new objects and their positions & Candidate object list:\newline \{name: bash, pos: (124,352)\} \\
\hline
\textbf{\textit{Object Verifier}} & Related screens $(S^{scr}_i, S^{scr}_{i+1})$ +\newline Object patch $O^v_{*}$ + $O^n_{*}$ \newline+ Existed $[O^n_0,O^n_1... ]$ & Verify image-description match and object novelty & Boolean: True \\
\hline
\textbf{\textit{Action Selector}} & Current screen $S^{scr}_i$ + Process $P_i$ \newline+ Candidate clusters $[A_{i_1}, A_{i_2}, ...]$ & Cluster selection based on current state & Action cluster ID: 5 \\
\hline
\textbf{\textit{Action Judge}} & Process $P_i$ + Action $a_i$ +\newline Related screens $(S^{scr}_i, S^{scr}_{i+1})$ & Evaluate consistency and effectiveness of action & Boolean pair: (True, False) \\
\hline
\textbf{\textit{Action Generator}} & Objects seq $[O_i, O_{i+1},..., O_n]$ +\newline Screen transitions$[S^{scr}_i, ..., S^{scr}_n]$ & Generate abstract action description & Action $a_i$: ``Attack the monster then return'' \\
\hline
\textbf{\textit{Action Coordinator}} & New action $a_i$ +\newline Current clusters $[A_0, A_1, ...]$ & Merge actions into existing action clusters & Selected cluster: \{id:4, name: battle with..., description: ...\} \\
\hline
\textbf{\textit{Progress Tracker}} & Process $P_i$ +\newline Screen pair $(S^{scr}_i, S^{scr}_j)$ & Update game stage transitions & New phase  $p_{i}$: ``Defeated the monster and obtained ...'' \\
\hline
\end{tabular}
\label{tab:vlm_roles}
\end{table*}

\subsection{Object Discovery Module}
In the \textbf{EPGA} framework, objects serve as the fundamental primitives for action construction. To ensure the agent can autonomously synthesize new actions and update its policy as the environment evolves, we design an object discovery module. This module supports grounding known objects, discovering novel interactive elements, and verifying their distinctiveness through visual matching and vision-language reasoning. By continuously identifying latent interactive elements, the agent can expand its action space and memory without manual supervision, ensuring the decision-making process remains dynamic and open-ended.

Let $S^{scr}_i$ denote the current game screen and $\mathcal{O} = \{O_k\}_{k=1}^N$ the set of known objects in the \textit{\textbf{Object Library}}, each associated with a reference image $O^v_k$. To identify previously seen objects, we perform rotation and scale-invariant template matching:
\begin{equation}
(x^*, y^*), s^* = \underset{\theta \in \Theta}{\arg\max}f_{S}(S^{scr}_i, R^\text{ro}(R^\text{re}(O^v_k, \alpha), \theta)),
\end{equation}
where $(x^*, y^*)$ is the location on the screen with the highest similarity score $s^*$, $f_{S}(\cdot,\cdot)$ denotes normalized cross-correlation, $R^\text{re}$ and $R^\text{ro}$ are resizing and rotation operations, $\Theta = \{-8^\circ, -4^\circ, 0^\circ, 4^\circ, 8^\circ\}$, and $\alpha = 1.0$. If $s^* > \tau$, the object is added to the visible set $\mathcal{O}_\text{exist}$. When visual changes significantly, game process advances, or no known objects remain, the VLM acts as an \textbf{\textit{Object Observer}}: 
\begin{equation}
f_\text{observe}(S^{scr}_i, P_i, \mathcal{O}_\text{exist}) \rightarrow \{(O^n_{{*}_j}, (x_j, y_j)),...\},
\end{equation}
where $P_i$ denotes current game process. This yields a list of candidate object names and their estimated screen coordinates. Each candidate point $(x_j, y_j)$ is probed via interaction. If the resulting state $S^{scr}_{i+1}$ differs from $S^{scr}_i$, the visual region around $(x_j, y_j)$ is cropped as $O^v_{*}$ and passed to the VLM acting as an \textbf{\textit{Object Verifier}}: 
\begin{equation}
f_\text{verify}(S^{scr}_i, S^{scr}_{i+1}, O^v_{*}, O^n_{{*}_j}, \{O^n_k\}_{k=1}^N) \rightarrow \text{Boolean}.
\end{equation}
Only if the verifier returns \texttt{True} does the agent proceed to compare with stored objects. 

To prevent redundant storage, the new patch $O^v_\text{*}$ is compared with existing object images $\{O^v_k\}_{k=1}^N$ using the structural similarity index (SSIM) \cite{bakurov2022structural}:
\begin{equation}
\text{SSIM}(O^v_\text{*}, O^v_k) = \frac{(2\mu_{*}\mu_k + C_1)(\sigma_{k*} + C_2)}{(\mu_{*}^2 + \mu_{*}^2 + C_1)(\sigma_{*}^2 + \sigma_k^2 + C_2)},
\end{equation}
where $\mu_{*}, \mu_k$ are means, $\sigma_{*}^2, \sigma_k^2$ are variances, and $\sigma_{k*}$ is the covariance of local pixel intensities. $C_1$ and $C_2$ are stability constants. If $\max_k~\text{SSIM}(O^v_\text{*}, O^v_k) < \delta$, the object is considered novel and added to memory $\mathcal{O}$. 

To ensure discovery efficiency and mitigate the impact of invalid interactions or state regressions, we implement a multi-stage verification and pruning strategy. If multiple interaction attempts fail or the probed object contradicts the VLM’s description, the agent logs the current state as an exploration failure. To accelerate learning, subsequent discovery triggers are immediately terminated if the game state remains stagnant and the agent attempts to re-probe nearby coordinates. 

Furthermore, to handle complex dependencies in strategy games where objects may not yield immediate state changes upon direct clicking, we introduce conditional interaction sequences. For instance, certain entities (e.g., hidden monsters) only become interactable after first activating a specific related card. Our module addresses this by combining known objects with candidate novel elements in joint interaction attempts, enabling the discovery of previously latent or "locked" entities. Conversely, to prevent redundant exploration, UI elements that trigger state changes but lack strategic depth (e.g., a "Return" button) are identified and excluded from future new-object discovery sequences once encountered. This hierarchical filtering ensures that the agent’s perception space expands only with meaningful, actionable objects while maintaining a robust trajectory during environmental evolution.

\subsection{Action Update Module}
To autonomously build and refine a repertoire of effective behaviors without human labeling or reward engineering, we design an action update module that supports the self-supervised discovery, selection, evaluation, and organization of actions. This module allows the agent to adaptively develop strategies through closed-loop interactions grounded in visual and semantic feedback. Rather than predicting actions from scratch, our agent reuses previously effective behavior clusters when appropriate. Given the current screen $S_t$, a visual encoder of pre-trained CLIP extracts its feature $S^{fea}_t$, which is matched via cosine similarity against stored cluster embeddings $\{S^{fea}_k\}_{k=1}^N$:
\begin{equation}
f_S(S^{fea}_t, S^{fea}_k) = \cos{(S^{fea}_t \cdot S^{fea}_k)}.
\end{equation}
Clusters above a similarity threshold are retained. The \textbf{\textit{Action Selector}} then incorporates both the screen $S^{scr}_i$ and interaction process $P_i$ to choose the most appropriate one: 
\begin{equation}
f_\text{select}(S^{scr}_i, P_i, [A_{i_1}, A_{i_2}, \dots]) \rightarrow A_k.
\end{equation}
If a valid cluster $A_k$ is selected and not suspended due to repeated failure, we perform tree-based action sampling using a variant of Upper Confidence Bound applied to Trees (UCT)\cite{couetoux2011continuous}:
\begin{equation}
\label{eq:uct}
\text{UCT}(a_i) = a_{f_i} + c\sqrt{(\ln A_{N_k})/a_{n_i}} + (1 - \text{Comp}(a_i)) \lambda,
\end{equation}
where $a_{f_i}$ is the fitness score stored in the \textit{\textbf{Action Library}}. This score quantifies the historical utility and success rate of action $a_i$ in the current state, serving as the primary component for exploitation and the main target for iterative policy updates. The second term, $c\sqrt{(\ln A_{N_k}) / a_{n_i}}$, follows the standard UCT formulation to encourage exploration of actions with fewer trials. Crucially, we introduce a feasibility-aware penalty, $(1 - \text{Comp}(a_i)) \lambda$, where $\text{Comp}(a_i) \in [0,1]$ represents object completeness—the alignment between the action's required objects and those detected in the current scene. By penalizing actions with missing or unrecognized constituents, this term suppresses invalid interactions and prioritizes executable actions, thereby accelerating policy convergence. Action probabilities are computed via softmax with temperature $T$:
\begin{equation}
a_{p_i} = \frac{\exp(\text{UCT}(a_i)/T)}{\sum_{a'} \exp(\text{UCT}(a')/T)}.
\end{equation}
Here, the \textit{explore} operation is also treated as an action, whose probability is computed in the same softmax distribution but without the fitness and completeness penalties. This balances the reuse of high-reward actions with exploration, while penalizing actions that are premature given current object availability. After executing $a_i$, the transition $(S^{scr}_i, S^{scr}_{i+1})$ is evaluated by the \textbf{\textit{Action Judge}} to determine both consistency (The state change is consistent with the action description.) and effectiveness (The action effectively advanced the game progress.): 
\begin{equation}
f_\text{judge}(P_i, (S^{scr}_i, S^{scr}_{i+1}), a_i) \rightarrow (\text{Boolean}, \text{Boolean}).
\end{equation}
If both flags are \texttt{True}, the fitness score $a_{f_i}$ is incremented by 1. When no suitable cluster matches the current state, previous attempts have failed, or the \textit{explore} operation is selected, the agent initiates exploration by prioritizing the action $a_i$ with the highest fitness score within the selected cluster (if no suitable cluster, initialize a new tree). The agent retrieves the corresponding tree and node from the \textit{\textbf{MCTS Library}}, attempting to append currently matched objects $\mathcal{O}_\text{exist}$ as new child nodes. The exploration proceeds by executing a sequence of operations on the related nodes. Once a sequence induces progress, a human-readable label $a_i$ is synthesized via the \textbf{\textit{Action Generator}}: 
\begin{equation}
f_\text{generate}([S^{scr}_i, S^{scr}_{i+1}, \dots, S^{scr}_n], [O_i, O_j, \dots, O_n]) \rightarrow a_i.
\end{equation}
This abstracted action $a_i$ is integrated into an appropriate cluster via the \textbf{\textit{Action Coordinator}}:
\begin{equation}
f_\text{coordinate}(a_i, [A_1, A_2, \dots, A_n]) \rightarrow A_j^{\text{updated}}.
\end{equation}
In this case, the MCTS structure is updated: the current tree is preserved, and the new action is anchored to the parent node that directly preceded the state change. 

Conversely, if all existing objects in $\mathcal{O}_\text{exist}$ fail to generate a valid action after being extended under the current MCTS branch, the cluster is temporarily suspended to avoid redundant search within the current context. This suspension remains in effect until a significant change in the game state is detected, at which point the cluster is reactivated for further exploration. The agent then exits the current cluster and restarts the selection process to identify a more promising cluster. At the conclusion of each action update cycle, we perform efficiency-driven pruning within the selected cluster. To optimize the search space, we identify and remove actions characterized by a high trial count ($a_{n_i}$) but a low fitness score ($a_{f_i}$). Through this iterative cycle of node expansion, conditional suspension, and strategic pruning, the action update module builds a grounded and high-utility action space from scratch while maintaining long-term computational efficiency.

\subsection{Game Process Tracking Module}
This module serves as a global observer that compensates for the local and myopic nature of object and action selection. In visually complex games, the agent may struggle to recognize when goals shift, sub-tasks complete, or new priorities emerge, especially under non-Markovian or partially observable dynamics. This module addresses that challenge by monitoring long-term state evolution and detecting semantically meaningful transitions in gameplay. To this end, we integrate a VLM as a \textbf{\textit{Progress Tracker}} that summarizes phase-level changes between screen states. It monitors the current process $P_i$, a reference state $S_{cp}$, and a current state $S_j$: 
\begin{equation}
f_\text{progress}(P_i, S_{cp}, S_j) \rightarrow p_i,   
\end{equation}
where $p_i$ is a natural language description of the inferred transition (e.g., ``a new wave of enemies has appeared''). Once a non-trivial change is detected, $p_i$ is appended to $P_i$ and $S_j$ becomes the new reference state for subsequent comparisons. This mechanism provides a lightweight, symbolic view of long-term game structure, abstracted from raw pixel dynamics. It allows upstream modules to make globally informed decisions, e.g., initiating object discovery when a new goal is detected, or resetting exploration when task stages shift. Compared to hand-coded heuristics, our learned process tracking generalizes across domains and enhances coordination between modules through a shared, evolving task history. The framework of \textbf{EPGA} are formalized in Algorithm~\ref{alg:epga-main}.

\begin{table}
\centering
\caption{Hardware and software configurations, where $t_a$, $t_o$, and $t_s$ are screen change ratio thresholds for action execution, object verification, and phased state change, respectively. }
\label{tab:hardware-hyperparams}
\begin{tabular}{p{4cm}|p{3.5cm}}
\hline
\textbf{Component} & \textbf{Specification / Value} \\ \hline
GPU for CLIP model & NVIDIA RTX1650\\
OS & Windows 10 \\
Python Version & 3.10 \\
Torch Version & 2.3.0+cu121 \\
Memory database & sqlite3 \\
CLIP Version & ViT-B/32 \\
VLM Model (via API) & GPT-4o, Claude3.7  \\
Image Resolution & 1280 $\times$ 720 \\
UCT Coefficient ($c$) & 5.0 \\
Cosine Similarity Threshold & 0.9 \\
Action Temperature ($T$) & 1.5 (initial) \\
Prompt Frequency & Once per session \\
change ratio threshold $t_a$ & 0.015 \\
change ratio threshold $t_o$ & 0.08\\
change ratio threshold $t_s$ & 0.3\\ \hline
\end{tabular}
\end{table}

\subsection{Efficiency and Key Actions Prioritization}
\label{sec:eff}
To ensure efficient learning and balanced development between the Object Discovery and Action Update modules, we regulate the ratio of candidate actions to active objects within a target range of 5–6. If this ratio exceeds the upper bound, the object discovery module is temporarily paused to prevent a combinatorial explosion during MCTS construction, which would slow exploration. Conversely, if the ratio falls below the lower bound, the Action Update module is deferred to ensure sufficient candidate actions per object, avoiding scenarios where some objects are under-explored. This dynamic throttling based on the action-to-object ratio balances exploration efficiency and memory growth, supporting stable and scalable learning. To emphasize high-value behaviors, we track actions associated with transitions between significant game phases. For each phase change, action IDs and names are recorded and processed by the VLM to identify critical actions that drive meaningful progress. These key actions receive augmented fitness scores, biasing future UCT sampling toward strategies more likely to yield substantive advancement. This mechanism enables the agent to systematically prioritize impactful behaviors while preserving self-supervised exploration across the full action-object space.

\begin{algorithm}[t]
\caption{Evolving of the Proposed EPGA framework.}
\label{alg:epga-main}
\KwIn{Raw screen $S_0$, initial process history $P_0$, thresholds $\{\delta, c, t_s, t_o, t_a\}$. Initialize VLM, CLIP encoder, and empty memory $\mathcal{M} \leftarrow \{\mathcal{O}, a, \mathcal{A}, \mathcal{T}\}$; $S_{cp} \leftarrow S_0, t \leftarrow 0$}.
\While{Game not terminated}{
    Capture current screen $S_t$, get visible set $\mathcal{O}_{exist}$\;
    \textbf{// Phase 1: Object Discovery \& Verification}\;
    \If{$diff(S_t, S_{cp}) > t_s$ \textbf{or} no objects detected \textbf{or} process updated \textbf{and} object-action-balance(sec: \ref{sec:eff})}{
        Detect candidates $\{(n_j, (x_j, y_j))\}$ via VLM \textit{Observer}\;
        \For{all candidate $j$}{
            Interact at $(x_j, y_j)$, capture $S_{t+1}$\;
            \If{$diff(S_{t+1}, S_t) > t_o$ \textbf{and} confirm novelty}{
                \If{$\max_{k} \text{SSIM}(O_{v_j}, O_{v_k}) < \delta$}{
                    Add object $(n_j, O_{v_j})$ to Object Library $\mathcal{O}$.
                }
            }
        }
    }
    \textbf{// Phase 2: Action Selection \& Update}\;
    Retrieve candidate action clusters via CLIP similarity\;
    \eIf{valid cluster $A_k$ found}{
        Sample $a_i \in A_k$ via Eq. \ref{eq:uct},
        execute $a_i$ (not Explore)\; 
        \If{Success}{
            $a_{f_i} \leftarrow a_{f_i} + 1$\;
        }
    }{
        Retrieve MCTS $\mathcal{T}_k$ and expand new node by $\mathcal{O}_{exist}$\;
        Execute exploratory from root to new node\; 
        \eIf{state change $> t_a$}{
            $a_{new} \leftarrow f_{generate}(S_t, S_{t+1}, \dots)$\; 
            Merge $a_{new}$ into Cluster $A_j$.
        }{
            Suspend current cluster until state $S_t$ transitions.
        }
        
    }
    \textbf{// Phase 3: Process Evolution \& Pruning}\;
    \If{$diff(S_{cp}, S_t) > t_s$ \textbf{and} high-level transition}{
        $P_t \leftarrow p_t \leftarrow f_\text{progress}(S_{cp}, S_t, P_t)$, $S_{cp} \leftarrow S_t$\; 
        Update fitness scores for key actions in $P_t$\;
        Prune actions where $a_{n_i}$ is high but $a_{f_i}$ is low. 
    }
    $t \leftarrow t + 1$.
}
\KwOut{Optimized policy and updated $\mathcal{M}$}    
\end{algorithm}

\section{Experiments}
We evaluate \textbf{EPGA} in two structurally diverse, open-ended environments: \textit{Slay the Spire}, which combines roguelike and deck-building mechanics, and \textit{Civilization V}, a hallmark 4X (eXplore, eXpand, eXploit, eXterminate) strategy game. Both games lack predefined APIs or explicit objectives, and present rich sets of interactive elements, partial observability, and long-term planning challenges. Our evaluation is structured around three questions through quantitative comparisons, ablation studies, and qualitative visualizations: 1) How does the agent perform compared to strong LLM-based baselines under identical conditions? 2) What is the impact of each individual module on the overall performance? 3) How does the agent’s internal memory evolve over time during autonomous exploration? Our aim is to assess whether the \textbf{EPGA} can effectively learn to play complex, previously unseen games from scratch—without any prior knowledge, reward shaping, or human intervention.

\textbf{Experimental Setup.} We conduct evaluations in two games under standardized conditions, i.e., \textit{Slay the Spire} (base difficulty: Ascension 0, character: Ironclad) and \textit{Civilization V} (Catherine/Russia civilization, map type: Earth, map size: Standard,  difficulty level: Prince, game pace: Standard). Our \textbf{EPGA} plays both games from raw visual inputs (RGB pixels) without internal API access. Zero human intervention is permitted post-launch. For semantic understanding and reasoning, we use GPT-4o as the VLM and CLIP ViT-B/32 as the visual encoder. The UCT-based coefficient $c$ is set to 5.0, and the cosine similarity threshold for action cluster retrieval is set to 0.9. Each configuration is evaluated across 5 independent runs, and all reported metrics are averaged over these trials. Table~\ref{tab:hardware-hyperparams} summarizes the hardware and software configurations required to run the agent, along with key hyperparameter settings.

\begin{table*}
\centering
\caption{Performance Comparison. Methods with * indicate \textbf{\textit{With Priors}} setting, while others are \textit{Zero Priors}. Progression is quantified by cleared floors in \textit{Slay the Spire} and survival turns in \textit{Civilization V}. In-game Score captures cumulative performance using official run scores (\textit{Slay the Spire}) and technologies unlocked (\textit{Civilization V}); Execution-Responsive Rate (ERR) reflects behavioral executability through the percentage of predicted actions yielding meaningful state transitions. Cost are average LLM tokens consumed per 100 steps, converted to USD for fair comparison across methods with varying episode lengths.}
\label{tab:main_results}
\setlength{\tabcolsep}{5pt}
\begin{tabular}{lcccccccc}
\toprule
\multirow{2}{*}{\textbf{Method}} & \multicolumn{4}{c}{\textbf{Slay the Spire}} & \multicolumn{4}{c}{\textbf{Civilization V}} \\
\cmidrule(lr){2-5} \cmidrule(lr){6-9}
 & Floors$\uparrow$ & Score$\uparrow$ & ERR (\%)$\uparrow$ & Cost$\downarrow$ & Turns$\uparrow$ & Techs.$\uparrow$ & ERR (\%)$\uparrow$ & Cost$\downarrow$ \\
\midrule
GPT4o*\cite{openai2024gpt4o} & 8.33 $\pm$ 0.58 & 48.73 $\pm$ 2.43 & 49.00 $\pm$ 2.00 & 1.01 $\pm$ 0.03 & 14.67 $\pm$ 2.08 & 1.00 $\pm$ 0.00 & 57.00 $\pm$ 1.00 & 0.91 $\pm$ 0.02 \\
Claude3.7*\cite{anthropic2025claude} & 1.00 $\pm$ 0.00 & 5.00 $\pm$ 0.00 & 79.00 $\pm$ 1.00 & 1.22 $\pm$ 0.04 & 19.33 $\pm$ 2.52 & 3.33 $\pm$ 0.58 & 92.00 $\pm$ 1.00 & 1.18 $\pm$ 0.09 \\
UITARS1.5*\cite{qin2025ui} & 1.00 $\pm$ 0.00 & 5.00 $\pm$ 0.00 & 54.00 $\pm$ 1.00 & 0.49 $\pm$ 0.24 & 11.33 $\pm$ 1.15 & 1.33 $\pm$ 0.58 & 92.00 $\pm$ 2.00 & 0.12 $\pm$ 0.01 \\
CRADLE*\cite{tan2024cradle} & 1.00 $\pm$ 0.00 & 5.00 $\pm$ 0.00 & - & 4.89 $\pm$ 0.23 & 0.00 $\pm$ 0.00 & 0.00 $\pm$ 0.00 & - & 4.23 $\pm$ 0.28 \\
B-U Agent*\cite{du2025rethinking} & 14.33 $\pm$ 1.15 & 91.33 $\pm$ 9.61 & 98.00 $\pm$ 1.00 & 2.68 $\pm$ 0.29 & 59.67 $\pm$ 8.74 & 8.66 $\pm$ 0.58 & 92.00 $\pm$ 1.00 & 3.55 $\pm$ 0.52 \\
SAG-Agent*\cite{tang2025experience} & 15.67 $\pm$ 0.47 & 104.00 $\pm$ 5.89 & 98.00 $\pm$ 1.00 & 2.15 $\pm$ 0.04 & 107.33 $\pm$ 6.55 & 13.33 $\pm$ 1.25 & 93.00 $\pm$ 1.00 & 3.03 $\pm$ 0.17 \\
\midrule
GPT4o\cite{openai2024gpt4o} & 1.00 $\pm$ 0.00 & 5.00 $\pm$ 0.00 & 72.00 $\pm$ 3.00 & 1.27 $\pm$ 0.09 & 6.33 $\pm$ 0.58 & 0.00 $\pm$ 0.00 & 37.00 $\pm$ 4.00 & 0.76 $\pm$ 0.03 \\
Claude3.7\cite{anthropic2025claude} & 1.00 $\pm$ 0.00 & 5.00 $\pm$ 0.00 & 28.00 $\pm$ 1.00 & 1.08 $\pm$ 0.06 & 0.00 $\pm$ 0.00 & 0.00 $\pm$ 0.00 & 15.00 $\pm$ 3.00 & 1.22 $\pm$ 0.13 \\
UITARS1.5\cite{qin2025ui} & 1.00 $\pm$ 0.00 & 5.00 $\pm$ 0.00 & 82.00 $\pm$ 1.00 & 0.09 $\pm$ 0.01 & 0.00 $\pm$ 0.00 & 0.00 $\pm$ 0.00 & 63.00 $\pm$ 1.00 & 0.10 $\pm$ 0.01 \\
\textbf{Ours} & \textbf{16.00 $\pm$ 0.67} & \textbf{109.00 $\pm$ 3.20} & \textbf{98.73 $\pm$ 1.00} & 2.93 $\pm$ 0.07 & \textbf{113.33 $\pm$ 9.75} & \textbf{15.33 $\pm$ 1.50} & \textbf{94.31 $\pm$ 1.00} & 3.81 $\pm$ 0.34 \\
\bottomrule
\end{tabular}
\end{table*}

\begin{table}
\setlength{\tabcolsep}{3pt}
\centering
\caption{
Ablation study results on \textit{Slay the Spire}, where \textbf{V} is Valid Proportion among the discovered objects (valid/stored), \textit{None} means no objects stored, \textit{Pre} is preloaded 30 basic objects, no new discovery.}
\label{tab:ablation_study}
\begin{tabular}{lcccc}
\toprule
Method & Floor$\uparrow$ & Score$\uparrow$ & ERR(\%)$\uparrow$ & V$\uparrow$ \\
\midrule
\textbf{Ours} & \textbf{16} & \textbf{109} & \textbf{98.7} & \textbf{89/100} \\
\quad \textit{w/o Memory} & 1 & 5 & 48.5 & None \\
\quad \textit{w/o Action Update} & 1 & 5 & 45.7 & 7/31 \\
\quad \textit{w/o Object Discovery} & 6 & 36 & 94.4 & Pre \\
\quad \textit{w/o Process Tracking} & 13 & 91 & 91.0 & 67/113 \\
\quad \textit{w/o Key Action Prior} & 10 & 67 & 77.3 & 74/86 \\
\bottomrule
\end{tabular}
\end{table}

\begin{table}
\centering
\caption{
Effect of the \textit{Action-Object Balance} mechanism on valid action growth across game rounds. 
$Run_i^A$ denotes the number of valid actions discovered in the $i$-th round. 
}
\label{tab:ablation_study2}
\begin{tabular}{lccc}
\toprule
Method & $Run_1^A$ & $Run_2^A$ & $Run_3^A$\\
\midrule
\textbf{Ours} & 47 & 124 & 212 \\
\quad \textit{w/o Action-Object Balance} & 45 & 80 & 119\\
\bottomrule
\end{tabular}
\end{table}

\subsection{Comparison with Peer Methods}
To the best of our knowledge, no existing agent framework supports fully autonomous gameplay from scratch, without predefined state representations, action spaces, or reward signals. Most prior works rely on structured APIs and task-specific prompts, making them unsuitable for our setting. The closest alternatives are UI-based agents, which interact with applications via visual elements, but still require substantial human priors. We compare the proposed \textbf{EPGA} against four state-of-the-art UI agents, i.e., GPT-4o \cite{openai2024gpt4o}, Claude 3.7 \cite{anthropic2025claude}, UI-TARS-1.5 \cite{qin2025ui}, CRADLE\cite{tan2024cradle}, Bottom-Up Agent \cite{du2025rethinking} and SAG-Agent \cite{tang2025experience}, under two distinct configurations: 1) \textbf{\textit{Zero Priors}}: The agent receives no prior knowledge—no object descriptions, action templates, or goals. This matches the initialization of our \textbf{EPGA}. 2) \textbf{\textit{With Priors}}: Agents are equipped with curated object lists and their functional descriptions, extracted from annotated screenshots. For \textit{Slay the Spire}, we provide 59 elements (e.g., monsters, cards, buttons); for \textit{Civilization V}, 46 examples (e.g., resources, units, UI buttons). 

As shown in Table~\ref{tab:main_results}, \textbf{EPGA} achieves dominant cross-game superiority under zero-prior and prior-assisted settings. In \textit{Slay the Spire}, \textbf{EPGA} clears 16 floors, surpassing the best prior-assisted baseline while achieving 109 in-game score with 98.73\% execution reliability. For \textit{Civilization V}, it attains 113-turn survival and 15 technologies researched at 94.31\% executability, establishing unprecedented advancement from raw pixel. Crucially, the performance gap highlights a fundamental architectural advantage of EPGA over existing UI-based frameworks, specifically the CRADLE\cite{tan2024cradle}, the Bottom-Up Agent \cite{du2025rethinking} and the SAG-Agent\cite{tang2025experience}. While they achieves relatively high execution rates by relying on a predefined library of interactable objects, it suffers from perceptual and strategic stagnation when faced with the procedural generation and high-density UI of Roguelike or strategy games. Both game environments evolve constantly, i.e., new cards, monsters, and UI panels appear that were not present in the initial state. Because Bottom-Up methods rely on static object-action mappings derived from predefined priors, they frequently "stall" in certain game sessions, repeatedly executing invalid actions when encountered elements deviate from their static library.

In contrast, EPGA treats objects as dynamic action primitives that are discovered and updated in real-time. By continuously identifying latent interactive elements and refining the MCTS-based action space, our framework maintains perceptual continuity and long-term trajectory tracking. This allows EPGA to adapt to novel game elements without any human-annotated knowledge, ensuring the exploration remains progressive even as the environment's complexity grows. As demonstrated in Table~\ref{tab:main_results}, while the SAG-Agent* reaches a bottleneck at 15 floors and 13 technologies despite having prior knowledge, EPGA consistently outperforms it under zero-prior conditions, proving that autonomous discovery is more robust than manual definition in highly stochastic environments. These results expose fundamental limitations in existing approaches: 1) \textit{Catastrophic prior dependency}: zero-prior baselines collapse completely (e.g., Claude3.7 without priors achieves 0 turns/technologies), while \textbf{EPGA} attains 113 turns/15 technologies without priors. 2) \textit{Action-strategy decoupling}: UITARS-1.5 with prior exhibits 92.00\% executability in \textit{Civilization V} yet progresses only 11 turns, revealing its inability to translate actions into strategic outcomes. 3) \textit{Generalization fragility}: prior-assisted baselines fail to match our zero-prior performance, confirming the scalability limits of manual engineering for complex strategy learning. 

\subsection{Ablation Study}
To assess the contribution of each module of \textbf{EPGA}, we conduct ablation experiments by systematically disabling specific components and measuring the resulting impact on agent performance under the same task configuration (as shown in Table~\ref{tab:ablation_study}). Specifically, we define four variants: 1) \textit{w/o Memory}: Disable long-term storage of discovered objects and actions, as well as action clustering. The agent interacts in a stateless manner, relying on random real-time exploration. 2) \textit{w/o Action Update}: Remove action effect validation and policy refinement. Actions are uniformly sampled from memory without regard to prior outcomes. 3) \textit{w/o Object Discovery}: Disable dynamic object discovery. The agent is initialized with a fixed set of 30 predefined basic interactive elements, but no new objects are added later. 4) \textit{w/o Game Process Tracking}: Prevent tracking of game-stage transitions. The agent relies solely on local visual cues for decision-making. 5) \textit{w/o Key Action Prior}: Remove the prioritization of high-value behaviors derived from phase-transition actions, causing the agent to lose strategic focus and exhibit less efficient progression despite stable exploration.
 
Disabling the memory module significantly degrades performance, as the agent loses the ability to accumulate prior knowledge and reuse successful strategies. Without the \textit{Action Update} module, action selection becomes purely stochastic, leading to poor policy refinement and frequent repetition of suboptimal behaviors. Omitting \textit{Object Discovery} restricts the agent to a static object set, hindering adaptability and causing stagnation in later stages where access to advanced game elements (e.g., new cards or potions) is critical. Without the \textit{Game Process Tracking} module, the agent lacks phase-level guidance, leading to redundant object discovery, degraded action validity, and ultimately poorer in-game performance. Lastly, removing the \textit{Key Action Prior} weakens strategic

prioritization—while the agent still explores effectively, it fails to emphasize phase-transition behaviors that drive long-term progress, leading to slower advancement and lower overall scores. Collectively, these results underscore the importance of each module of \textbf{EPGA} framework in facilitating robust, self-supervised learning and sustainable gameplay progression. As shown in Table~\ref{tab:ablation_study2}, as exploration rounds progress the agent encounters more scenes and thus more new objects; without a balanced update between objects and actions, this influx of objects significantly slows down action learning, since MCTS construction and action exploration become increasingly burdened by the growing object set.

\subsection{Visualization}
\begin{figure}[t]
    \centering
    \includegraphics[width=\linewidth]{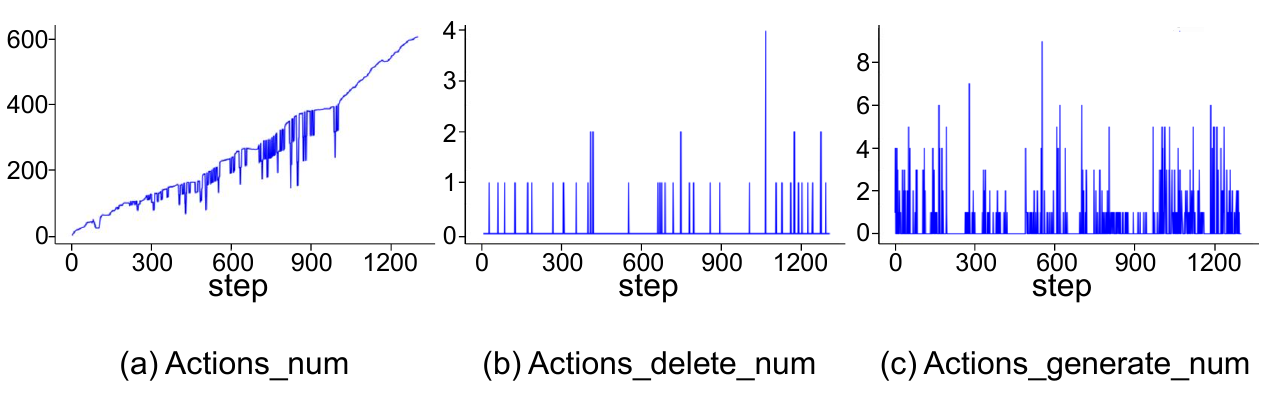} 
    \caption{\textbf{Action Evolution in \textit{Slay the Spire}.} (a) Cumulative actions stored. (b) Invalid actions removed per step. (c) Newly generated actions per step. These reflect continual policy expansion and pruning.}
    \label{fig:action_evolution}
\end{figure}

\begin{figure}[t]
    \centering
    \includegraphics[width=\linewidth]{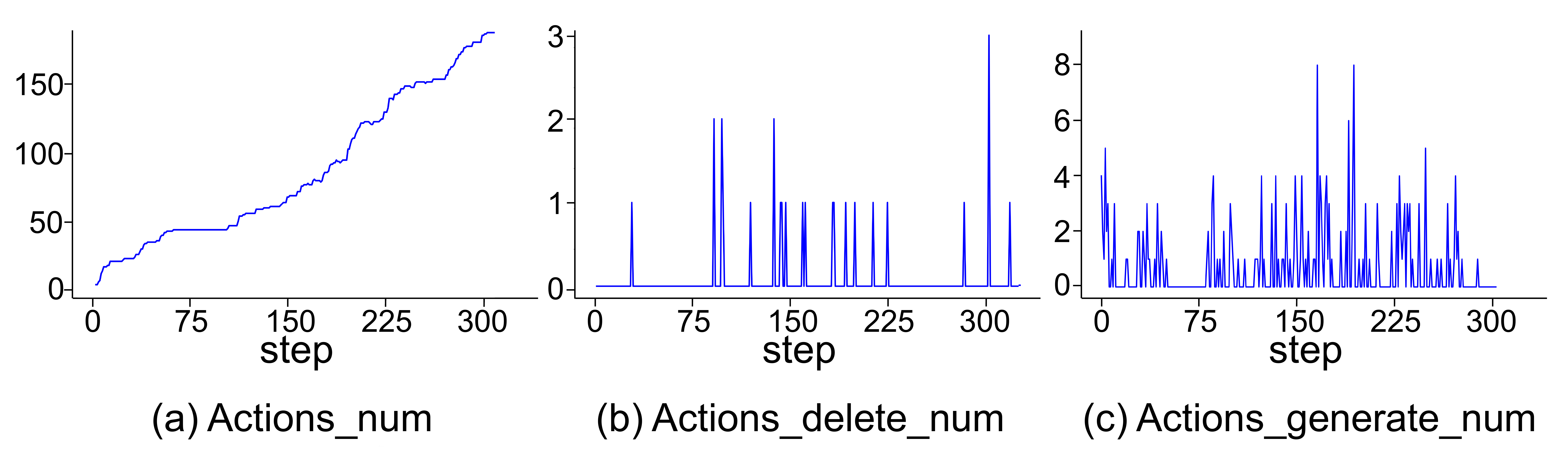} 
    \caption{\textbf{Action Evolution in \textit{Civilization V}.} (a) Cumulative actions stored. (b) Invalid actions removed per step. (c) Newly generated actions per step.}
    \label{fig:action_evolution_Civilization_V}
\end{figure}

\begin{figure}[h]
    \centering
    \includegraphics[width=\linewidth]{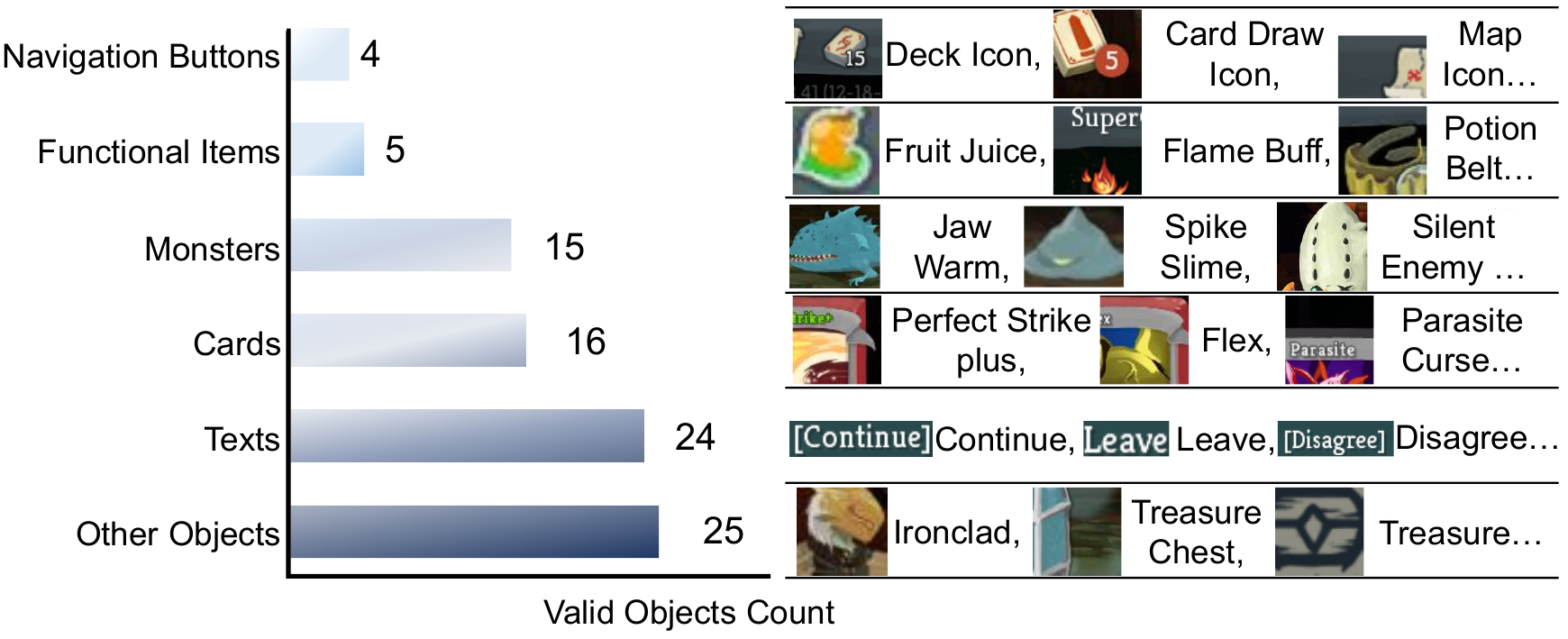} 
    \caption{\textbf{Discovered Objects after a Full Round of Exploration in \textit{Slay the Spire}.} For analysis purposes, objects are manually grouped into categories: \textit{Monsters} (stage bosses), \textit{Cards} (new cards obtained), \textit{Texts} (interactive text buttons), \textit{Functional Items} (potions, relics), \textit{Navigation Buttons} (UI buttons like map), and \textit{Other Objects} (map icons, etc.).}
    \label{fig:object_discovery}
\end{figure}

\begin{figure}[h]
    \centering
    \includegraphics[width=\linewidth]{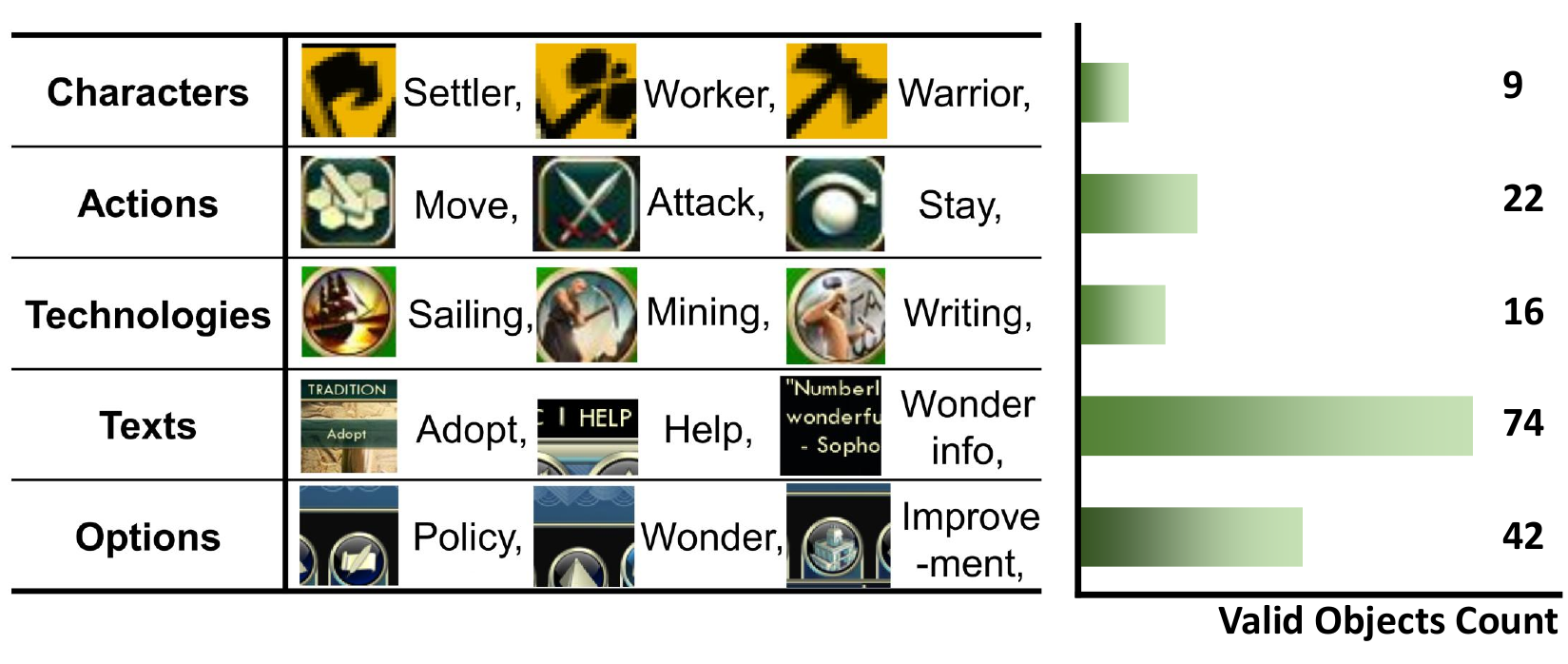} 
    \caption{\textbf{Discovered Objects after a Full Round of Exploration in \textit{Civilization V}.} Objects are grouped into five categories: \textit{Characters} (map units and leaders), \textit{Actions} (current-turn operations), \textit{Technologies} (unlockable techs), \textit{Texts} (interactable text elements), and \textit{Options} (menu and UI entries).}
    \label{fig:object_discovery_Civilization_V}
\end{figure}

\begin{figure}[t]
    \centering
    \includegraphics[width=\linewidth]{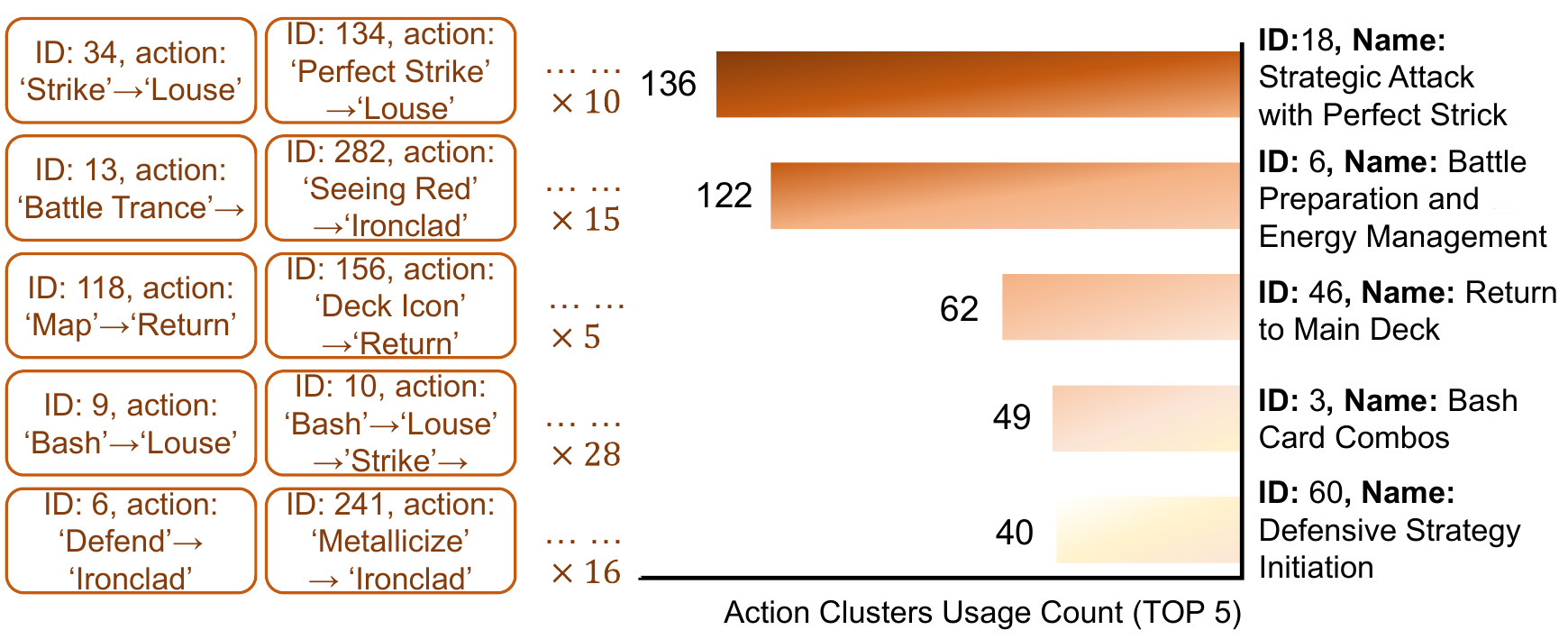} 
    \caption{\textbf{Frequent Action Clusters in \textit{Slay the Spire}.} The top 5 most-invoked clusters after one full exploration episode. Each row shows(left to right): representative actions, number of actions in the cluster, total invocation count, and a short semantic description.}
    \label{fig:cluster_examples}
\end{figure}

To demonstrate how the proposed \textbf{EPGA} constructs knowledge through environmental interaction, we visualize three core learning dimensions in \textit{Slay the Spire} and \textit{Civilization V}): action space evolution, object discovery, and strategic action clustering. Figure~\ref{fig:action_evolution} and Figure~\ref{fig:action_evolution_Civilization_V} reveals continuous policy refinement: cumulative actions grow to 600+ (Figure~\ref{fig:action_evolution}(a)-\textit{Slay the Spire}-1200 step) and 150+ (Figure~\ref{fig:action_evolution}(a)-\textit{Civilization V}-300 step) through active generation (1-8 new actions/step, Figure~\ref{fig:action_evolution}(c) and Figure~\ref{fig:action_evolution_Civilization_V}(c)) while maintaining quality via pruning (1-2 invalid actions/step, Figure~\ref{fig:action_evolution}(b) and Figure~\ref{fig:action_evolution_Civilization_V}(b)). The object discovery matrix (Figure~\ref{fig:object_discovery} and Figure~\ref{fig:object_discovery_Civilization_V}) confirms comprehensive environment understanding with 80+ semantically categorized entities \textit{Slay the Spire} and 160+ semantically categorized entities in \textit{Civilization V}) enabling hierarchical feature abstraction. Most critically, Figure~\ref{fig:cluster_examples} demonstrates autonomous strategy formation: agent-clustered actions reveal emergent gameplay patterns like offensive sequences, defensive maneuvers, and navigation protocols, validating unsupervised learning of transferable tactics from raw experience. Table~\ref{tab:explore_top10} summarizes the most frequently explored action clusters in \textit{Civilization V}, further demonstrating the emergence of high-level strategic behaviors and the growing understanding of the game's underlying mechanics.


\begin{table*}[t]
\centering
\caption{\textbf{Top 10 Action Clusters Ranked by Explore Count in \textit{Civilization V}.}}
\label{tab:explore_top10}
\begin{tabular}
{>{\centering\arraybackslash}m{0.5cm}>{\centering\arraybackslash}m{4cm}>{\raggedright\arraybackslash}m{10cm}>{\centering\arraybackslash}m{0.5cm}}
\toprule
\multicolumn{1}{c}{\textbf{ID}} &
\multicolumn{1}{c}{\textbf{Name}} &
\multicolumn{1}{c}{\textbf{Description}} &
\multicolumn{1}{c}{\textbf{Count}} \\
\midrule
8  & Technology Exploration & Interact with the technology tree to view detailed information and decide on strategic advancements. & 26 \\
\midrule
31 & Navigating Civ 5's Civilopedia & This sequence involves navigating through the Civilopedia in Civilization V, including clicks in ‘units\_home\_page’ and ‘buildings\_tab’ to access various topics. Useful for learning mechanics. \textbf{Precaution}: Focus on relevant sections to avoid overload. & 24\\
\midrule
9  & CIVILOPEDIA Navigation & Navigate the in-game encyclopedia to explore terrain, buildings, and policies. Helps in understanding strategy. Ensure accurate navigation to minimize confusion. & 20 \\
\midrule
27 & Open Social Policies & Opens the Social Policies menu to choose civilization-wide bonuses affecting culture, economy, and strength. \textbf{Precaution}: Changes are impactful—choose wisely. & 19 \\
\midrule
24 & Search and Navigate in Civilopedia & Use search to explore topics like terrain or units. \textbf{Precautions}: (1) Use accurate terms. (2) Understand hierarchy for efficient navigation. & 16\\
\midrule
1  & Unit Orders and Strategic Management & Manage unit turns, attack enemy units, and execute full strategies. \textbf{Precautions}: (1) Check battle odds. (2) Don’t skip unit actions. & 15 \\
\midrule
14 & Game Turn Advancement & Focuses on advancing turns only after validating prior actions. Ensures readiness for future turns. & 15 \\
\midrule
26 & Social Policies Management & Navigate and manage policies in the Civilopedia, including backtracking and information lookup. & 15\\
\midrule
11 & Next Turn Confirmation & Advance to next turn after selecting enemy unit. & 14\\
\midrule
29 & Social Policies Navigation & Explore social policy menu and return to main UI. & 13\\
\bottomrule
\end{tabular}
\end{table*}

\section{Case Study}
\begin{figure*}[h]
    \centering
    \includegraphics[width=0.95\linewidth]{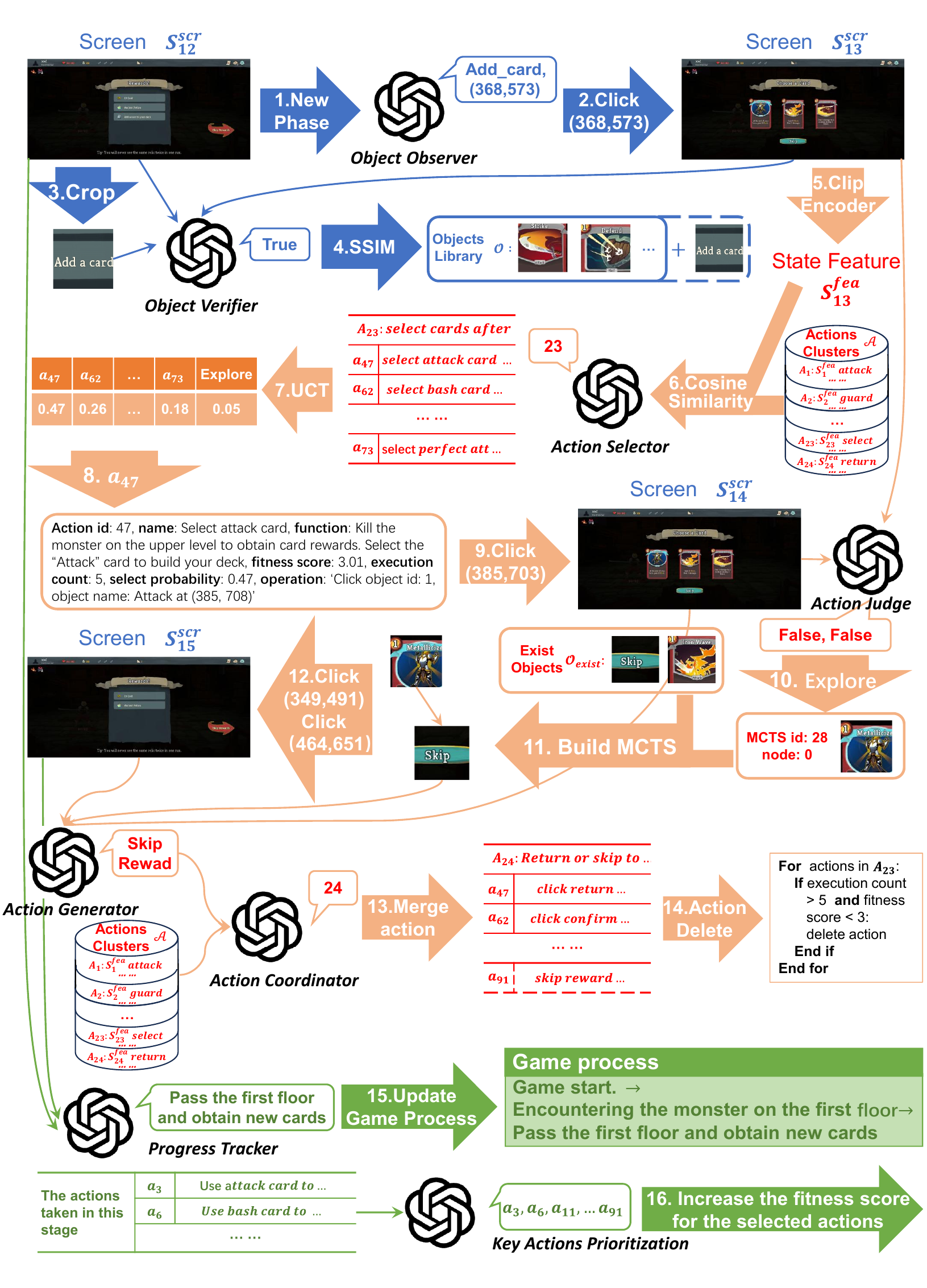} 
    \caption{Case Study of a Full Exploration-and-Learning Cycle of our \textbf{Evolving Pixel-Based Game Agent}.}
    \label{fig:case_study}
\end{figure*}

Figure~\ref{fig:case_study} illustrates a complete exploration-and-learning cycle of our \textit{EPGA}. The agent first detects newly appeared objects and verifies their validity and novelty through LLM reasoning and SSIM checks before adding them to the object library (Steps 1--4). The current screen is then encoded into a CLIP feature to retrieve relevant action clusters, from which the LLM selects an action intention and a UCT-based strategy samples an executable action whose fitness is updated according to the resulting visual changes (Steps 5--9). If no valid interaction is found, the agent enters exploration mode, expands the corresponding MCTS by composing existing objects until a meaningful interaction is discovered, and uses the LLM to name, cluster, and optimize the newly learned action while pruning redundant ones (Steps 10--14). Finally, successful stage completion updates the Game Process Module and reinforces key actions that contributed to the success (Steps 15--16). Through this closed loop of exploration, reasoning, and self-optimization, \textit{EPGA} continuously refines its object and action libraries without external supervision.

\begin{figure*}[h]
    \centering
    \includegraphics[width=\linewidth]{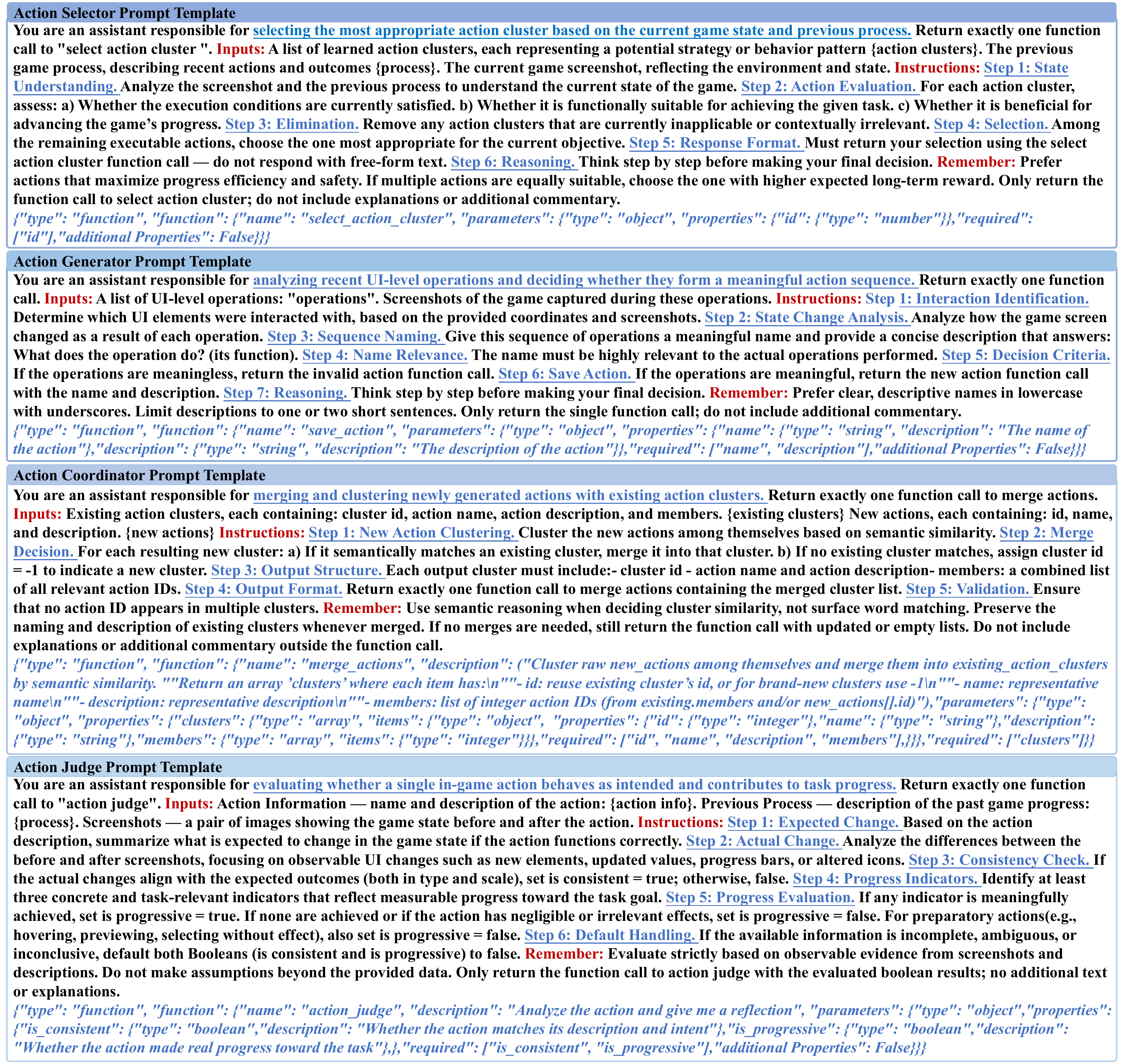} 
    \caption{\textbf{General structure of prompts used in Action modules.}}
    \label{fig:action_prompt}
\end{figure*}

\begin{figure*}[h]
    \centering
    \includegraphics[width=\linewidth]{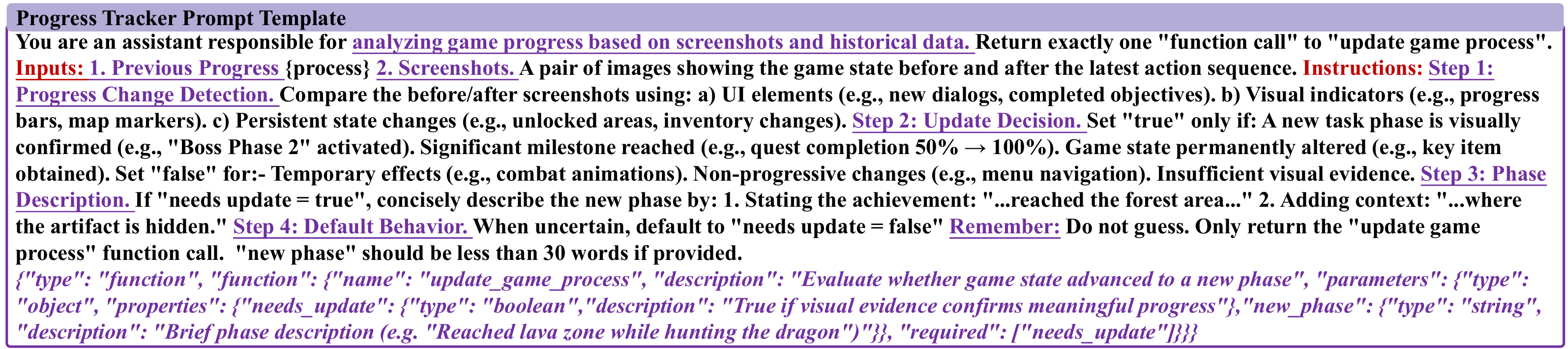} 
    \caption{\textbf{General structure of prompts used in Progress modules.}}
    \label{fig:progress_prompt}
\end{figure*}

\begin{figure*}[t]
    \centering
    \includegraphics[width=\linewidth]{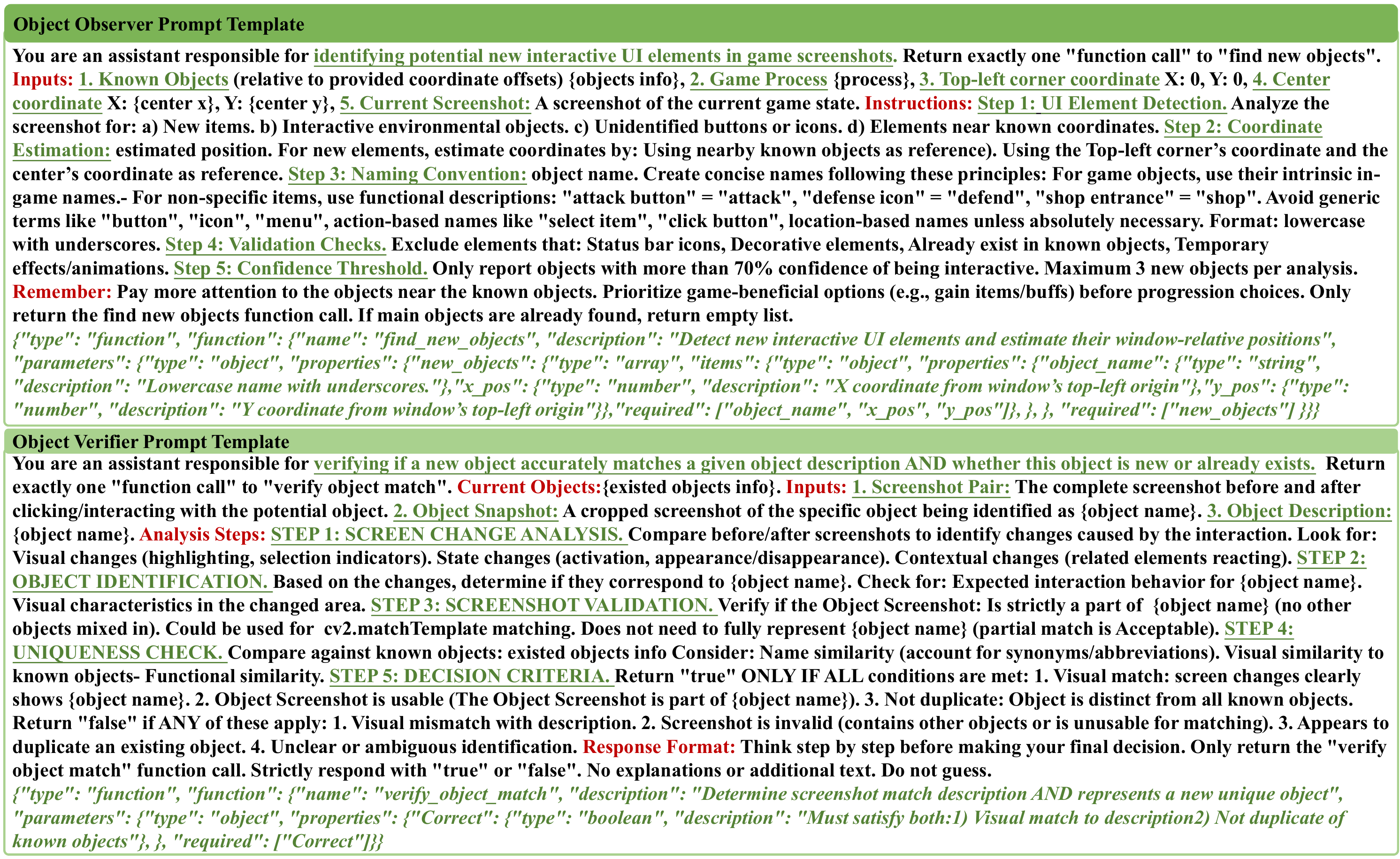} 
    \caption{\textbf{General structure of prompts used in Object modules.}}
    \label{fig:object_prompt}
\end{figure*}

As shown in Figure~\ref{fig:action_prompt}, Figure~\ref{fig:progress_prompt}, and Figure~\ref{fig:object_prompt}, we list the general structure of prompts used in different modules.

\clearpage
\newpage

\clearpage
\newpage

\section{Conclusion}

We presented \textbf{EPGA}, an evolving pixel-based agent for autonomous learning in complex and dynamically evolving game environments directly from raw visual observations. By continuously discovering interactive elements, constructing and refining actions, maintaining structured memory, and tracking long-horizon game processes, \textbf{EPGA} progressively builds its own interaction knowledge without predefined states, actions, rewards, or manually collected object representations. This design reduces the dependence on task-specific environmental engineering and allows the agent's perception, actions, and strategies to evolve with the environment. Experiments on two structurally different environments, \textit{Slay the Spire} and \textit{Civilization V}, demonstrate sustained gameplay progression, high execution reliability, and competitive performance against strong multimodal and UI-based agents, including methods equipped with manually prepared object knowledge. \textbf{EPGA} reaches 16 floors in \textit{Slay the Spire} and 15 technologies in \textit{Civilization V}, surpassing the strongest prior-assisted baselines without pre-collected object screenshots or annotations. Ablation studies further verify the contributions of object discovery, action updating, structured memory, and process tracking to long-term progression, while trajectory analyses show how reusable interaction knowledge is progressively accumulated during exploration. These results suggest that autonomous construction and continual refinement of interaction knowledge provide a scalable alternative to static environment-specific engineering for multimodal agents. Despite these advances, exploration efficiency remains constrained by the growing interaction space, and the current framework lacks explicit retrospective or counterfactual reasoning. Future work will investigate adaptive exploration scheduling and counterfactual reasoning to improve exploration efficiency and strategy refinement, with the broader goal of developing multimodal agents that can continually acquire and reorganize their own knowledge in previously unseen and evolving environments.

\bibliography{main}
\bibliographystyle{IEEEtran}

\end{document}